\documentclass{article}

\usepackage[final]{nips_2018}

\usepackage{amsmath}
\usepackage{hyperref}
\usepackage{url}
\usepackage{graphicx}
\usepackage{subcaption}

\usepackage[utf8]{inputenc} 
\usepackage[T1]{fontenc}    
\usepackage{hyperref}       
\usepackage{url}            
\usepackage{booktabs}       
\usepackage{amsfonts}       
\usepackage{nicefrac}       
\usepackage{microtype}      

\title{State representation learning with recurrent capsule networks}

\author{Louis Annabi, Michael Garcia Ortiz \\
AI Lab\\
Softbank Robotics Europe\\
Paris, France \\
\texttt{louis.annabi@gmail.com} \\
\texttt{mgarciaortiz@softbankrobotics.com}
}

\begin{document}

\maketitle

\begin{abstract}
  Unsupervised learning of compact and relevant state representations has been proved very useful at solving complex reinforcement learning tasks \cite{DBLP:journals/corr/abs-1803-10122}. In this paper, we propose a recurrent capsule network \cite{10.1007/978-3-642-21735-7_6} that learns such representations by trying to predict the future observations in an agent's trajectory.
\end{abstract}

\section{Introduction}

We study the mechanisms that could allow task-independent and self-supervised learning for embodied agents. Specifically, we focus here on the acquisition of perceptive skills, in particular representation for physical objects.

Recently, capsule networks were proposed as a model for representing visual objects. A capsule is a set of artificial neurons that returns both an activation and instantiation parameters. The activation is the estimated probability of the feature being present in the input image, and is similar to the outputs of neurons in convolutional neural networks. The instantiation parameters are estimations of different properties of the detected feature, such as position, orientation, lighting or color. However, reaching configurations where each capsule models one visual entity can be tricky without using a classification loss, here we are looking for an approach that doesn't rely on external information.

Sensorimotor prediction is often proposed to learn representations. It has been extensively used in reinforcement learning literature as an auxiliary task to learn both state representations and forward models. In \cite{DBLP:journals/corr/abs-1803-10122}, the authors first learn a latent representation using variational auto-encoders \cite{DBLP:journals/corr/KingmaW13} and then learn a forward model by trying to predict future latent states. In \cite{merlin}, the authors propose a fully differentiable RL architecture trained, specially, with a reconstruction loss of the inputs and a KL divergence between the estimated and actual latent state distributions.

In this work, we explore the use of capsule networks for state representation learning. We reformulate the principle of capsules in order to use them in the framework of sensorimotor 
prediction. This allows us to build a state representation preserving the benefits explained in \cite{10.1007/978-3-642-21735-7_6} while avoiding the use of external data.

\section{Model}

\begin{figure}[ht]
\begin{center}
\includegraphics[width=0.55\textwidth]{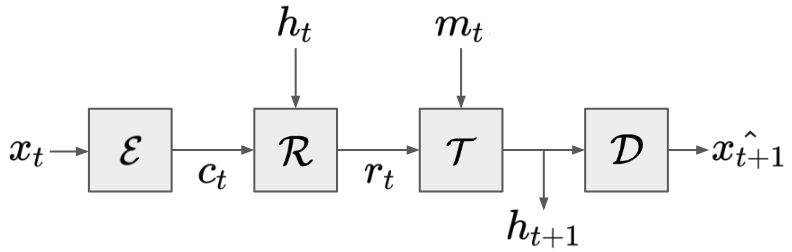}
\end{center}
\caption{Recurrent model used to train capsules. $\mathcal{E}$ encodes the observation $x_t$ into a capsule representation $c_t$, which is fed to a recurrent cell $\mathcal{R}$ that produces the state capsule representation $r_t$. $\mathcal{T}$ uses $r_t$ together with the motor commands $m_t$ to predict the next state $h_{t+1}$. Finally, the decoder $\mathcal{D}$ estimates the next observation $\hat{x_{t+1}}$}
\label{fig:model}
\end{figure}

Our predictive model is made of four functional blocks, that are repetitively used at each time step : an encoder, a recurrent cell, a transformation cell and a decoder. We use artificial neural networks for each of these blocks, making the model fully differentiable. Figure \ref{fig:model} shows how those blocks are organized to form the full model in charge of predicting the future observations.

\subsection{Encoder}

\begin{figure}[ht]
\begin{center}
\includegraphics[width=0.7\textwidth]{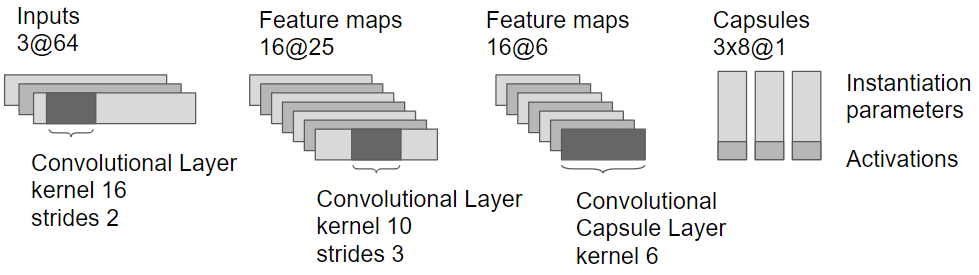}
\end{center}
\caption{Architecture of the capsule network used for encoding the observations. The network is composed of two convolutional layers, and one convolutional capsule layer.}
\label{fig:caps_encoder}
\end{figure}

The encoder outputs a capsule representation $c_t$ of shape ($k$, $d+1$) where $k$ is the number of capsules, $d$ is the dimension of the instantiation parameter vector returned by each capsule, and the $+1$ stands for the additional activation output for each capsule. Figure \ref{fig:caps_encoder} shows the architecture of the encoder network that we used in our experiments.

\subsection{Custom recurrent cell}

The recurrent cell is a custom neural network model that updates the state capsule representation $h_t$ with the capsule representation of the observation $c_t$. This network outputs an updated state capsule representation $r_t$. All three variables have the same shape ($k$, $d+1$). The custom recurrent cell is making use of the structure of the capsule output (activation dimension denoted $a$ and instantiation parameters dimensions denoted $v$) and performs better than traditional recurrent cells (GRU, LSTM) on the sensorimotor prediction task. Here is the detail of the operation performed:

\begin{equation*}
\begin{aligned}
r_t^a &= \max (h_t^a, c_t^a) \\
\epsilon_t &= f(h_t^a, c_t^a) = \frac{1}{1 + \frac{h_t^a}{c_t^a}^2} \\
r_t^v &= (1 - \epsilon_t) \cdot  h_t^v + \epsilon_t \cdot g(h_t^v, c_t^v; w)
\end{aligned}
\end{equation*}

Where $g$ is a linear function of parameter $w$ taking as input the concatenation of $h_t^v$ and $c_t^v$. The main idea here is to modulate the output with an update parameters $\epsilon_t$ depending on the activations of the state capsule representation and the activations coming from the observation capsule representation.

\subsection{Transformation cell}

The transformation cell applies the transformation induced by the continuous action $m_t$ to the state representation $r_t$. We split the instantiation parameters in two categories : the "fixed" parameters denoted $vf$ that will not be affected by the transformation, and the "variable" parameters denoted $vv$ that are allowed to change. The transformation cell first computes a matrix representation $z_t$ for the action $m_t$ using a multi-layer perceptron. The action representation is a square matrix of size $d_v < d$, where $d_v$ is the number of variable parameters. Then the capsules received in input are transformed by multiplying the vector of variable parameters with the action matrix representation. Here is the detail of the operation performed :

\begin{equation*}
\begin{aligned}
h_{t+1}^a &= r_t^{a} \\
h_{t+1}^{vf} &= r_t^{vf} \\
h_{t+1}^{vv} &= \text{matmul}(\text{mlp}(m_t) , r_t^{vv})
\end{aligned}
\end{equation*}

\subsection{Decoder}

The decoder is responsible for computing an estimation of the next observation $\hat{x}_{t+1}$ based on the transformed state $h_{t+1}$. It is composed of a deep transposed convolutional neural network shared for each capsule, outputting an individual reconstruction for each capsule, and a custom merge layer, in charge of computing the observation estimation based on the individual reconstructions.

\section{Experiment}

\subsection{Environment}

To validate our approach, we tested our model in a partially observable environment with first-person observations. To be able to conduct many experiments, we chose to use the 2D simulator Flatland \cite{casellesdupr2018flatland}, where we simulate an environment similar to \cite{Eslami1204}. Having a 2D simulator produces observations in 1D, making it faster to conduct experiments. Our environment is composed of three simple round objects of three different colors, as shown in figure \ref{fig:env}. An agent equipped with vision sensor (raw RGB input) moves by random longitudinal displacement, lateral displacement, and rotation.

\subsection{Results}

The model is trained with a mean squared error loss on the estimated observations. Two additional losses are used to ensure a proper convergence : a sparsity constraint on the activations in the hidden state, and a mean squared error between the hidden state in input of the recurrent cell and the output of the recurrent cell. 

\begin{figure}[ht!]
\centering
\begin{subfigure}{0.29\textwidth}
\includegraphics[width=\textwidth]{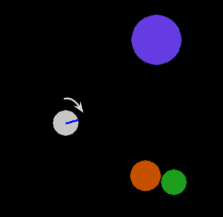}
\caption{One of the simulated environments, with the agent in light gray and the three objects in purple, orange and green. The position and size of the three objects vary along episodes.}
\label{fig:env}
\end{subfigure}
\hspace{0.3cm}
\begin{subfigure}{0.67\textwidth}
\includegraphics[width=\textwidth]{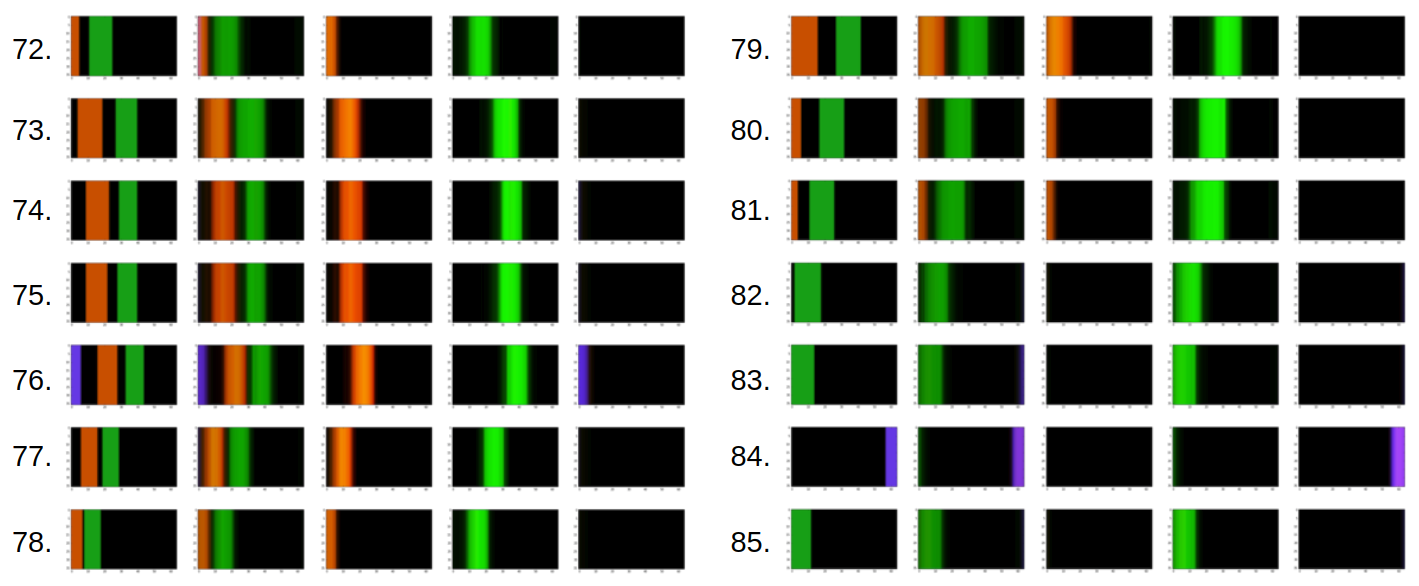}
\caption{Next observation prediction at the end of training. This figures shows a sample from a trajectory of length 100 on which the model was not trained. The five columns correspond to: 1. The real observation. 2. The predicted observation. 3. The individual prediction of the first capsule. 4. The individual prediction of the second capsule. 5. The individual prediction of the third capsule.}
\label{fig:seq}
\end{subfigure}
\end{figure}

Figure \ref{fig:seq} shows the sensorimotor prediction at the end of training. We can see from this figure that each capsule represents one object of the environment, which was the desired configuration. We can also notice that the purple object is properly predicted in frame 84 even though it had left the field of vision of the agent since frame 77, which indicates that our model is able to properly track the position of an object through long sequences of actions.

\begin{figure}[ht]
\begin{center}
\includegraphics[width=\textwidth]{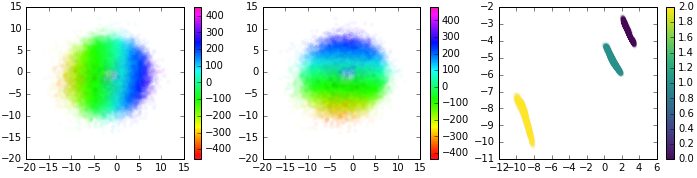}
\end{center}
\caption{Individual capsule representation plotted according to two dimensions of the variable parameters (left and middle), and the two dimensions of the fixed parameters (right). The color map used correspond to the real relative X (left) and Y (middle) coordinates of the encoded object in the agent's relative coordinate system, and to the color of the object (right).}
\label{fig:representation}
\end{figure}

Moreover, when looking at the learned representations of each object (see figure \ref{fig:representation}), we observe a high correlation between some of the capsule dimensions with the real relative position of the corresponding object with regard to the agent. This suggests that the agent has build a notion of space through this prediction process, without any information directly related to space in its observations (such as depth).

\section{Future work}

First, we need to run more experiments on the learned representations. Early results tend to show that this capsule representation performs well as input of a reinforcement learning algorithm in a navigation task, compared to learning from raw images. To further verify the quality of our model for reinforcement learning, we would need to compare it with other state representation learning methods (like \cite{DBLP:journals/corr/abs-1803-10122}) and on several environments.

In the future, we would like to extend this model, for instance by adding routing with a second layer of capsules, and apply it to more complex environments, with more objects and more variability between objects.

\bibliographystyle{ACM-Reference-Format}
\bibliography{caps}

\end{document}